\documentclass{article}

\usepackage{microtype}
\usepackage[accepted]{icml2020}
\usepackage{algorithm}
\usepackage{algpseudocode}

\usepackage{titlesec}
\newcommand{\addperiod}[1]{#1.}
\titleformat{\paragraph}[runin]{\normalfont\bfseries\color{black}\setlength{\parindent}{0pt}}{\theparagraph}{0.25em}{\addperiod}

\usepackage{subdepth}
\usepackage{amsfonts,bbm,bm,courier}
\usepackage{setspace,relsize}
\usepackage{graphicx,xcolor,placeins}
\usepackage{enumitem}
\usepackage{hyperref}
\usepackage{mathtools}
\hypersetup{%
  colorlinks=true,%
  urlcolor=blue,%
  linkcolor=blue,%
  citecolor=blue,%
  linkbordercolor=blue,%
  pdfborderstyle={/S/U/W 1}%
}

\usepackage{array,booktabs,tabularx,multirow,colortbl,hhline}
\newcommand{\cell}[2]{\setlength{\tabcolsep}{0pt}\begin{tabular}{#1}#2 \end{tabular}}
\newcommand{\bfcell}[2]{\setlength{\tabcolsep}{0pt}\bfseries\begin{tabular}{#1}#2\end{tabular}}
\newcommand{\sccell}[2]{\setlength{\tabcolsep}{0pt}\scshape\begin{tabular}{#1}#2\end{tabular}}

\usepackage{eqnarray,mathtools}
\usepackage[oldsyntax]{stackengine}

\DeclareMathOperator*{\argmin}{argmin}

\newtheorem{definition}{Definition}
\newtheorem{proposition}{Proposition}

\setitemize{leftmargin=1em,parsep=2pt,label=\raisebox{0.25ex}{\tiny$\bullet$}}
\setlist[enumerate]{leftmargin=*, label= {\arabic*.}, itemsep=0.5em}
\newlist{thmlist}{enumerate}{1}
\setlist[thmlist]{leftmargin=*,label=\raisebox{0.25ex}{\tiny$\bullet$}, topsep=0.2em,itemsep=2pt}

\newcommand{\textds}[1]{{\footnotesize\texttt{#1}}}
\newcommand{\textfn}[1]{{\textit{#1}}}
\newcommand{\textcaption}[1]{{\textcolor{gray}\footnotesize\textsf{#1}}}

\newcommand{\indic}[1]{\mathbbm{1}[#1]}

\newcommand{\txplus}[0]{+}
\newcommand{\txminus}[0]{-}

\newcommand{\R}{\mathbb{R}}

\newcommand{\wb}{\bm{w}}
\newcommand{\xb}{\bm{x}}
\newcommand{\coefs}[0]{\wb}
\newcommand{\coef}[1]{w_{#1}}
\newcommand{\poscoef}[1]{\coef{#1}^\txplus{}}
\newcommand{\negcoef}[1]{\coef{#1}^\txminus{}}

\newcommand{\st}{\textnormal{s.t.}}
\newcommand{\miprange}[3]{{#1}={#2},...,{#3}}

\newcommand{\emin}[1]{\epsilon^\textnormal{lb}}
\newcommand{\emax}[1]{\epsilon^\textnormal{ub}}

\newcommand{\marginsym}[0]{\gamma}
\newcommand{\errorsym}{l}

\newcommand{\error}[1]{{\errorsym_{#1}}}

\newcommand{\conflictset}[0]{K}

\algnewcommand\algorithmicinput{\textbf{Input}}
\renewcommand\algorithmicensure{\textbf{Output:}}
\algnewcommand\algorithmicinitialize{\textbf{Initialize}}
\algnewcommand\algorithmicbigstep{\textbf{Step}}
\algnewcommand\INPUT{\item[\algorithmicinput]}
\algnewcommand\INITIALIZE{\item[\algorithmicinitialize]}
\algnewcommand{\STEP}[1]{\item[\algorithmicbigstep]{\textbf{#1}}}
\algnewcommand{\InputExplanation}[2][.6\linewidth]{\leavevmode\hfill\makebox[#1][r]{~{\footnotesize{#2}}}}
\algnewcommand{\InitializationExplanation}[2][.6\linewidth]{\leavevmode\hfill\makebox[#1][r]{~{\footnotesize{#2}}}}
\algnewcommand{\alginput}[2]{\INPUT{#1}\InputExplanation{#2}}
\algnewcommand{\StateComment}[2]{\State{#1}\InputExplanation{#2}}
\algnewcommand{\alginitialize}[2]{\Statex{#1}\InitializationExplanation{#2}}
\algrenewcommand\algorithmiccomment[2][]{#1\hfill\textit{\scriptsize{#2}}}

\newcommand{\Hset}[0]{\mathcal{H}}

\newcommand{\emprisk}[1]{\hat{R}({#1})}

\newcommand{\baseclf}[0]{h_0}
\newcommand{\diffclf}[1]{g_{#1}}
\newcommand{\altclf}[1]{g_{#1}}

\newcommand{\cfun}[2]{\delta(#1,#2)}

\newcommand{\epsset}[1]{S_\epsilon({#1})}
\newcommand{\setConflicts}{\alpha_\epsilon}
\newcommand{\singleConflicts}{\delta_\epsilon}

\newcommand{\gitrepo}[0]{https://github.com/charliemarx/pmtools}

\begin{document}

\twocolumn[
\icmltitle{Predictive Multiplicity in Classification}
\begin{icmlauthorlist}
\icmlauthor{Charles T. Marx}{hvf}
\icmlauthor{Flavio du Pin Calmon}{seas}
\icmlauthor{Berk Ustun}{ucsd}
\end{icmlauthorlist}

\icmlaffiliation{hvf}{Haverford College}
\icmlaffiliation{seas}{Harvard SEAS}
\icmlaffiliation{ucsd}{UC San Diego}

\icmlcorrespondingauthor{Charles T. Marx}{cmarx@haverford.edu}
\icmlcorrespondingauthor{Berk Ustun}{berk@ucsd.edu}

\icmlkeywords{multiplicity, classification, integer programming, recidivism prediction, measurement, contestability}

\vskip 0.3in
]

\printAffiliationsAndNotice{}
\begin{abstract}
Prediction problems often admit competing models that perform almost equally well. This effect challenges key assumptions in machine learning when competing models assign conflicting predictions. In this paper, we define \emph{predictive multiplicity} as the ability of a prediction problem to admit competing models with conflicting predictions. We introduce formal measures to evaluate the severity of predictive multiplicity and develop integer programming tools to compute them exactly for linear classification problems. We apply our tools to measure predictive multiplicity in recidivism prediction problems. Our results show that real-world datasets may admit competing models that assign wildly conflicting predictions, and motivate the need to measure and report predictive multiplicity in model development.
\end{abstract}

\section{Introduction}
\label{Sec::Introduction}

Machine learning algorithms are often designed to fit a best model from data. For example, modern methods for empirical risk minimization fit a model by optimizing a specific objective (e.g., error rate) over models that obey a specific set of constraints (e.g., linear classifiers with equal TPR between groups). In an ideal scenario where stakeholders agree on such a problem formulation~\citep{passi2019problem} and we are given a large dataset of representative examples, the use of machine learning may still lead to ethical challenges if there are multiple best-fitting models.

In machine learning, \emph{multiplicity} refers to the ability of a prediction problem to admit multiple \emph{competing models} that perform almost equally well. Several works mention that prediction problems can exhibit multiplicity~\citep[see e.g.,][]{mountain1989combined,mccullagh1989generalized}, but few discuss its implications. The work of \citet{breiman2001statistical} is a major exception. In a seminal position paper, Breiman describes how multiplicity challenges explanations derived from a single predictive model: \emph{if one can fit multiple competing models -- each of which provides a different explanation of the data-generating process -- how can we tell which explanation is correct?}

Drawing parallels between the discordant explanations of competing models and the discordant testimonies of witnesses in the motion picture ``Rashomon," Breiman refers to this dilemma the \emph{Rashomon effect}. In the context of his work, the Rashomon effect is -- in fact -- an argument against the misuse of explanations. Seeing how prediction problems can exhibit multiplicity, we should not use the explanations of a single model to draw conclusions about the broader data-generating process, at least until we can rule out multiplicity.

Machine learning has changed drastically since Breiman coined the Rashomon effect. Many models are now exclusively built for prediction~\citep{kleinberg2015prediction}. In applications like lending and recidivism prediction, predictions affect people~\citep{binns2018s}, and multiplicity raises new challenges when competing models assign conflicting predictions. Consider the following examples:
\begin{itemize}[leftmargin=0pt,label={}]

\item \emph{Recidivism Prediction}: Say that a recidivism prediction problem admits competing models with conflicting predictions. In this case, a person who is predicted to recidivate by one model may be predicted not to recidivate by a competing model that performs equally well. If so, we may want to ignore predictions for this person or even forgo deployment.

\item \emph{Lending}: Consider explaining the prediction of a loan approval model to an applicant who is denied a loan~\citep[e.g., by producing a counterfactual explanation for the prediction][]{martens2014explaining}. If competing models assign conflicting predictions, then these predictions may lead to contradictory explanations. In this case, reporting evidence of competing models with conflicting predictions would mitigate unwarranted rationalization of the model resulting from \emph{fairwashing}~\citep{aivodji2019fairwashing,laugel2019dangers,slack2020fooling} or \emph{explanation bias}~\citep[][]{koehler1991explanation}.

\end{itemize}

In this work, we define \emph{predictive multiplicity} as the ability of a prediction problem to admit competing models that assign conflicting predictions. Predictive multiplicity affects key tasks in modern machine learning -- from model selection to model validation to post-hoc explanation. In such tasks, presenting stakeholders with information about predictive multiplicity empowers them to challenge these decisions.

Our goal is to allow stakeholders to measure and report predictive multiplicity in the same way that we measure and report test error. To this end, we introduce formal measures of predictive multiplicity in classification:
\begin{itemize}[leftmargin=0pt, label={}]

\item \emph{Ambiguity}: How many individuals are assigned conflicting predictions by any competing model?

\item \emph{Discrepancy}: What is the maximum number of predictions that could change if we were to switch the model that we deploy with a competing model?

\end{itemize}
Both measures are designed to support stakeholder participation in model development and deployment (see e.g., Figure \ref{Fig::ToyExampleXOR}). For example, ambiguity counts the number of individuals whose predictions are determined by the decision to deploy one model over another. These individuals should have a say in model selection and should be able to contest the predictions assigned to them by a model in deployment.

The main contributions of this paper are as follows:
\begin{enumerate}
    
\item We introduce formal measures of predictive multiplicity for classification problems: \emph{ambiguity} and \emph{discrepancy}. 
    
\item We develop integer programming tools to compute ambiguity and discrepancy for linear classification problems. Our tools compute these measure \emph{exactly} by solving non-convex empirical risk minimization problems over the set of competing models. 
    
\item We present an empirical study of predictive multiplicity in recidivism prediction. Our results show that real-world datasets can admit competing models with highly conflicting predictions, and illustrate how reporting predictive multiplicity can inform stakeholders in such cases. For example, in the ProPublica COMPAS dataset \citep[][]{angwin2016machine}, we find that a competing model that is only 1\% less accurate than the most accurate model assigns conflicting predictions to over 17\% of individuals, and that the predictions of 44\% of individuals are affected by model choice. %

\end{enumerate}

\begin{figure}[t]
\centering
\resizebox{0.8\linewidth}{!}{
\begin{tabular}{ccc*{4}c}
\multicolumn{1}{c}{\cell{c}{\textcaption{Feature}\\\textcaption{Values}}} &
\multicolumn{2}{c}{\cell{l}{\textcaption{\# Data Points}\\\textcaption{Where $y_i = \pm 1$}}} &
\multicolumn{4}{c}{\cell{c}{\textcaption{Predictions of Best}\\\textcaption{Linear Classifiers}}}
\vspace{0.5em}\\
$(x_1, x_2)$&  
$n^+$&
$n^-$&
$\hat{h}_a$&
$\hat{h}_b$&
$\hat{h}_c$&
$\hat{h}_d$\\
\cmidrule(lr){1-1}  
\cmidrule(lr){2-3}  
\cmidrule(lr){4-7}
$(0, 0)$ & 
$0$&
$25$&
$-$&
$-$&
$-$&
\multicolumn{1}{>{\columncolor{red!25}[0pt]}c}{$+$}\\ 
$(0, 1)$& 
$25$&
$0$&
$+$&
$+$&
\multicolumn{1}{>{\columncolor{red!25}[0pt]}c}{$-$}&
$+$\\ 
$(1, 0)$& 
$25$&
$0$&
$+$&
\multicolumn{1}{>{\columncolor{red!25}[0pt]}c}{$-$}&
$+$&
$+$\\
$(1, 1)$&
$0$&
$25$& 
\multicolumn{1}{>{\columncolor{red!25}[0pt]}c}{$+$}&
$-$&
$-$&
$-$\\
\cmidrule(lr){1-1} 
\cmidrule(lr){2-3} 
\cmidrule(lr){4-7}
\end{tabular}
}
\caption{Classifiers with conflicting predictions can perform equally well. We show 4 linear classifiers that optimize accuracy on a 2D classification problem with 100 points. The predictions of any 2 models differ on 50 points. Thus, discrepancy is 50\%. The predictions of 100 points vary based on model choice. Thus, ambiguity is 100\%.} %
\label{Fig::ToyExampleXOR}
\end{figure}

\subsection{Related Work}
\label{Sec::RelatedWork}

\paragraph{Multiplicity}

\begin{figure}[tb]
\centering
\begin{tabular}{cc}
    \cell{c}{\textcaption{Rashomon Effect}} & \cell{c}{\textcaption{Predictive Multiplicity}} \\
    \includegraphics[width=0.4\linewidth]{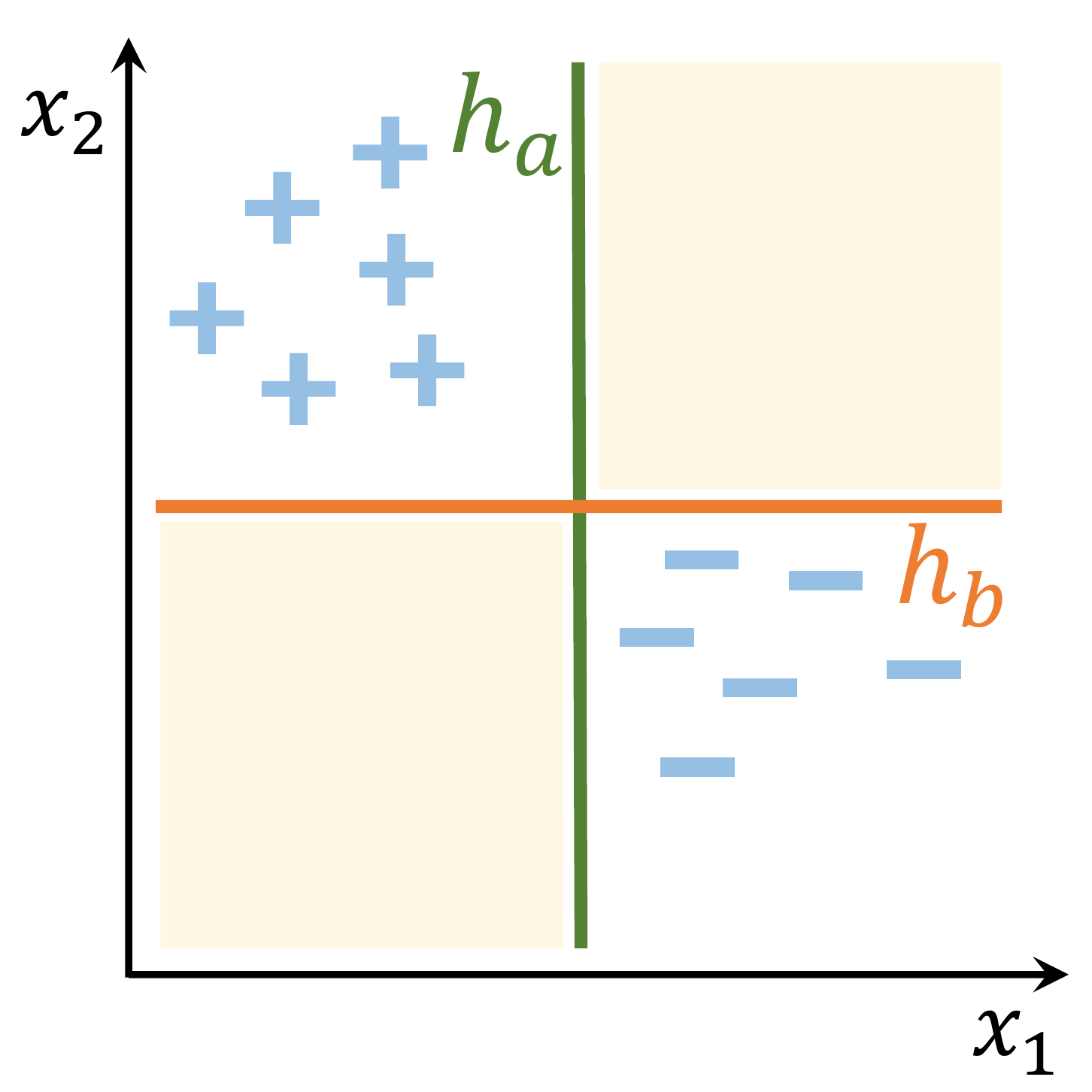} & \includegraphics[width=0.4\linewidth]{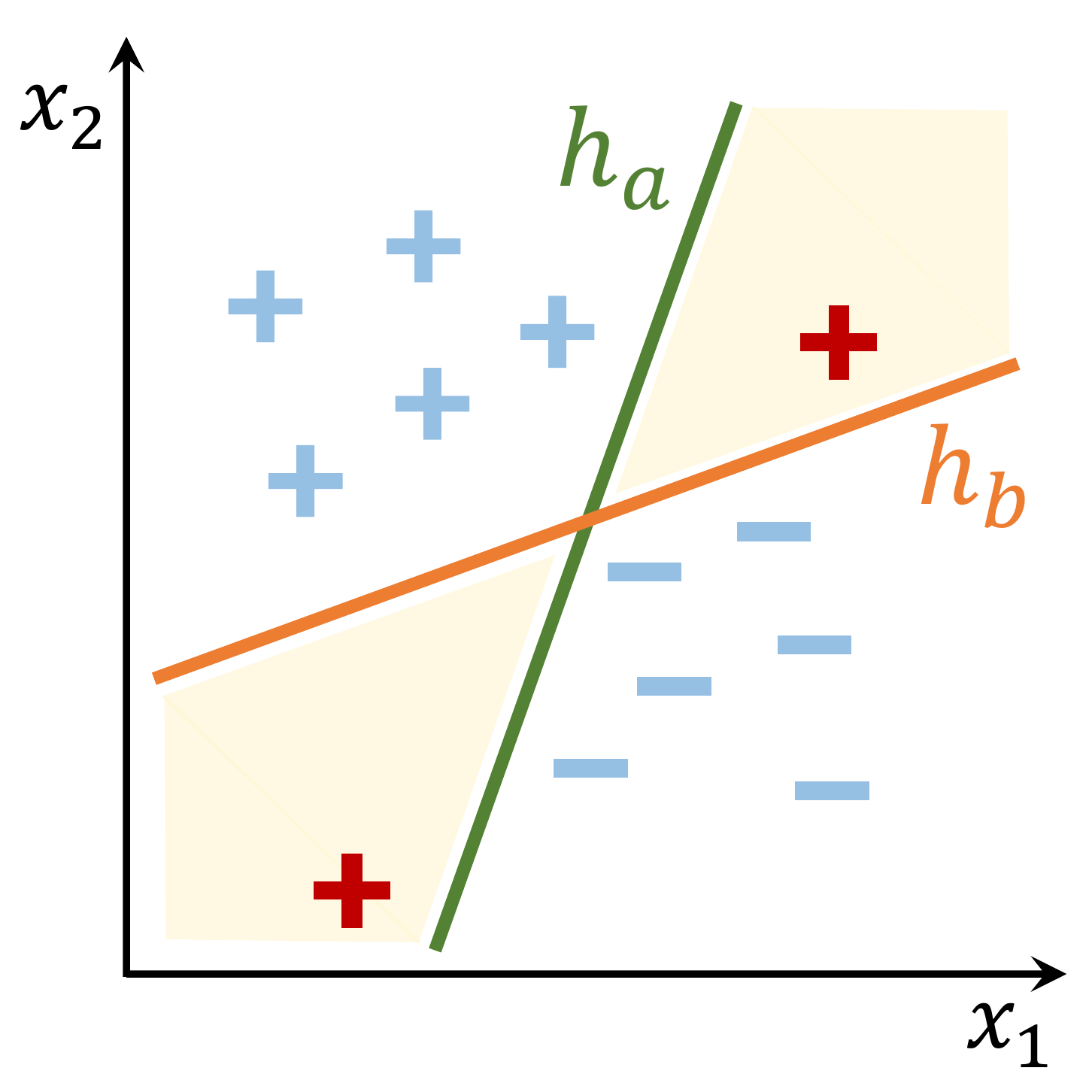} \\
\end{tabular}
\caption{Predictive multiplicity reflects irreconcilable differences between the predictions of competing models. Here, we depict two classification problems where the competing classifiers $h_a$ and $h_b$ optimize accuracy. We highlight points that are assigned conflicted predictions in red and regions of conflict in yellow. On the left, $h_a$ and $h_b$ assign the same predictions on the training data but produce conflicting explanations of the importance of $x_1$ vs. $x_2$, as per the Rashomon effect. On the right, $h_a$ and $h_b$ assign conflicting predictions on the training data as per predictive multiplicity.}
\label{Fig::TypesOfMultiplicity}
\end{figure}

Recent work in machine learning tackles multiplicity from the ``Rashomon" perspective. \citet[][]{fisher2018all} and \citet{dong2019variable} develop methods to measure variable importance over the set of competing models. \citet{semenova2019study} present a formal measure of the size of the set of competing models and use it to characterize settings where simple models perform well. Our work differs from this stream of research in that we study competing models \emph{with conflicting predictions} (see Figure \ref{Fig::TypesOfMultiplicity}). Predictive multiplicity reflects irreconcilable differences between subsets of predictions -- similar to the impossibility results in fair machine learning literature~\citep[][]{chouldechova2017fair,kleinberg2016inherent,corbett2017algorithmic}. 

\paragraph{Model Selection}

Techniques to resolve multiplicity can be broadly categorized as approaches for tie-breaking and reconciliation. Classical approaches for model selection break ties using measures like AIC, BIC, or K-CV error~\citep[see e.g.,][]{mcallister2007model, ding2018model}. These approaches are designed to improve out-of-sample performance. However, they may fail to do so when problems exhibit predictive multiplicity. In Figure \ref{Fig::ToyExampleXOR} for example, tie-breaking would not improve out-of-sample performance as all competing models perform equally well.

\paragraph{Bayesian approaches} 

Bayesian approaches explicitly represent multiplicity through posterior distributions over models. Posterior distributions are commonly used to construct a single model for deployment via majority vote or randomization procedures~\citep[see e.g.,][]{mcallester1999some,germain2016pac}. In theory, however, posterior distributions could allow for an ad hoc analyses of predictive multiplicity -- e.g., by counting conflicting predictions over a set of models sampled from the posterior \citep[see][]{dusenberry2020analyzing}. While valuable, these analyses may underestimate the severity of predictive multiplicity because the sample would not contain all competing models. 

\paragraph{Integer Programming}

Our work is part of a recent stream of research on integer programming methods for classification~\citep{nguyen2013algorithms,belotti2016handling,ustun2015slim, ustun2019actionable,ustun2019fairness}. We present methods to compute measures of predictive multiplicity for linear classification problems by solving integer programs. Integer programming allows us to count conflicting predictions over the full set of competing classifiers – i.e., all models that attain $\epsilon$-optimal values of a discrete performance metric like the error rate. In contrast, traditional approaches for reducing computation would produce unreliable estimates of predictive multiplicity. For example, if we were to count conflicting predictions over models that attain $\epsilon$-optimal values of a convex surrogate loss. In this case, we could underestimate or overestimate predictive multiplicity because models that attain near-optimal performance may differ significantly from models that attain near-optimal values of a surrogate loss.

\section{Framework}
\label{Sec::ProblemStatement}

In this section, we introduce measures of predictive multiplicity. For clarity of exposition, we present measures for binary classification problems. Our measures generalize to problems where models optimize other performance metrics (e.g., AUC), predict multiple outcomes, or obey additional constraints on performance or model form.

\paragraph{Preliminaries}

We start with a dataset of $n$ examples $\{(\xb_i, y_i)\}_{i=1}^{n}$ where each example consists of a feature vector $\xb_i = (1,x_{i1},\ldots,x_{id}) \in \R^{d+1}$ and a label $y_i \in \{\pm 1\}$. We use the dataset to fit a \emph{baseline classifier} $h:\R^{d+1} \to \{\pm 1\}$ from a hypothesis class $\Hset$ by minimizing empirical risk (i.e., training error):%
\begin{align*}
    \baseclf{} \in \argmin_{h \in \Hset} \emprisk{h}
\end{align*}
where $\emprisk{h} := \frac{1}{n} \sum_{i=1}^n \indic{h(\xb_i) \neq y_i}.$ 

This practice is aligned with the goal of optimizing true risk (i.e., test error) when $\baseclf{}$ generalizes. Generalization is a reasonable assumption in our setting as we work with a simple hypothesis class (see e.g., empirical results in Table \ref{Table::BaselineComparisons}). Fitting models that optimize performance on all of the training data is a best practice in machine learning~\citep[see e.g.,][for a discussion]{cawley2010over}. 
\footnote{For example, in a typical setting where we need to control overfitting by tuning hyperparameters over a validation dataset, we would first find hyperparameters that optimize an estimate of out-of-sample error (e.g., mean 5-CV error). We would then fit a model to optimize performance for these hyperparameters using all of the training data.}

\paragraph{Competing Models}

We measure predictive multiplicity over a set of classifiers that perform almost as well as the baseline classifier. We refer to this set as the \emph{$\epsilon$-level set} and to $\epsilon$ as the \emph{error tolerance}.
\begin{definition}[$\epsilon$-Level Set]
Given a baseline classifier $\baseclf{}$ and a hypothesis class $\Hset$, the \emph{$\epsilon$-level set} around $\baseclf{}$ is the set of all classifiers $h \in \Hset$ with an error rate of at most $\emprisk{\baseclf{}} + \epsilon$ on the training data: $$\epsset{\baseclf{}} := \{h \in \Hset: \emprisk{h} \leq \emprisk{\baseclf{}} +  \epsilon\}.$$
\end{definition}

Predictive multiplicity can arise over an $\epsilon$-level set where $\epsilon = 0$ (see e.g., Figure \ref{Fig::ToyExampleXOR}). Despite this, we typically measure predictive multiplicity over an $\epsilon$-level set where $\epsilon > 0$. This is because a competing model with near-optimal performance on the training data may outperform the optimal model in deployment. In such cases, it would not be defensible to rule out competing models due to small differences in training error.

In practice, $\epsilon$ should be set so that the $\epsilon$-level set is likely to include a model that attains optimal performance in deployment. This can be achieved by computing confidence intervals for out-of-sample performance (e.g., via bootstrapping or cross-validation) or by using generalization bounds (e.g., by setting $\epsilon$ so that with high probability the $\epsilon$-level set contains the model that optimizes true risk).

\paragraph{Predictive Multiplicity}

A prediction problem exhibits \emph{predictive multiplicity} if competing models assign conflicting predictions over the training data.
\begin{definition}[Predictive Multiplicity]
Given a baseline classifier $\baseclf{}$ and an error tolerance $\epsilon$, a prediction problem exhibits predictive multiplicity over the $\epsilon$-level set $\epsset{\baseclf{}}$ if there exists a model $h \in \epsset{\baseclf{}}$ such that $h(\xb_i) \neq \baseclf{}(\xb_i)$ for some $\xb_i$ in the training dataset.
\end{definition}
The fact that competing models assign conflicting predictions means that model selection will involve arbitrating irreconcilable predictions. 

In what follows, we present formal measures of predictive multiplicity. Each measure evaluates the severity of predictive multiplicity by counting the number of examples that are assigned conflicting predictions by competing models in the $\epsilon$-level set. 

\begin{definition}[Ambiguity]
\label{Def::Ambiguity}
The ambiguity of a prediction problem over the $\epsilon$-level set $\epsset{\baseclf{}}$ is the proportion of points in a training dataset that can be assigned a conflicting prediction by a competing classifier $h \in \epsset{\baseclf{}}$:
$$\setConflicts(\baseclf{}) := \frac{1}{n} \sum_{i=1}^n \max_{h \in \epsset{\baseclf{}}} \indic{h(\xb_i) \neq \baseclf{}(\xb_i)}.$$
\end{definition}

\begin{definition}[Discrepancy]
\label{Def::MaximumDiscrepancy} 
The discrepancy of a prediction problem over the $\epsilon$-level set $\epsset{\baseclf{}}$ is the maximum proportion of conflicting predictions between the baseline classifier $\baseclf{}$ and a competing classifier $h \in \epsset{\baseclf{}}$:
$$\singleConflicts(\baseclf{}) := \max_{h \in \epsset{\baseclf{}}} \frac{1}{n} \sum_{i=1}^n \indic{h(\xb_i) \neq \baseclf{}(\xb_i)}.$$
\end{definition}

Ambiguity represents the number of predictions that can change over the set of competing models. This reflects the number of individuals whose predictions are determined by model choice, who could contest the prediction assigned to them by the deployed model, and who should have a say in model selection. 

Discrepancy represents the maximum number of predictions that can change if we switch the baseline classifier with a competing classifier. This reflects that in practice, in order to change multiple predictions, the conflicting predictions must all be realized by a single competing model.

We end with a discussion of the relationship between accuracy and predictive multiplicity. In Proposition \ref{Prop::maxConflictsBound}, we bound the number of conflicts between the optimal model and a model in the $\epsilon$-level set. We include a proof in Appendix \ref{Appendix::Proofs}. 
\begin{proposition}[Bound on Discrepancy]
\label{Prop::maxConflictsBound}
The discrepancy between $\baseclf{}$ and any competing classifier in the $\epsilon$-level set $h \in \epsset{\baseclf{}}$ obeys: $$ \singleConflicts \leq 2\emprisk{\baseclf{}} + \epsilon.$$
\end{proposition}
Proposition \ref{Prop::maxConflictsBound} demonstrates how the severity of predictive multiplicity depends on the accuracy of a baseline model. Specifically, a less accurate baseline model provides more ``room" for predictive multiplicity. This result motivates why it is important to measure discrepancy and ambiguity using the best possible baseline model.

\newpage
\section{Methodology}
\label{Sec::Methodology}

In this section, we present integer programming tools to compute ambiguity and discrepancy for linear classification problems.

\subsection{Overview}

\paragraph{Baseline Classifier}

Our tools compute ambiguity and discrepancy given a baseline linear classifier $\baseclf{}$ -- i.e., the classifier that we would typically deploy. In our experiments, we use a baseline classifier $\baseclf{}$ that minimizes the error rate, which we fit using a MIP formulation in Appendix \ref{Appendix::MIP}. This ensures that multiplicity does not arise due to suboptimality. Thus, the only way to avoid multiplicity is to change the prediction problem – i.e., by changing the dataset, the model class, or the constraints. %

\paragraph{Path Algorithms}

We present \emph{path algorithms} to compute ambiguity and discrepancy for all possible $\epsilon$-level sets. Path algorithms efficiently compute the information needed to show how ambiguity and discrepancy change with respect to $\epsilon$ (see Figure \ref{Fig::ErrorProfiles}). These plots relax the need for practitioners to choose $\epsilon$ a priori, and calibrates their choice of $\epsilon$ in settings where small changes in $\epsilon$ may produce large changes in ambiguity and discrepancy.

\paragraph{MIP Formulations}

We compute ambiguity and discrepancy by fitting classifiers from the $\epsilon$-level set. We fit each classifier by solving a discrete empirical risk minimization problem. We formulate each problem as a \emph{mixed integer program} (MIP). Our MIP formulations can easily be changed to compute predictive multiplicity for more complex prediction problems -- e.g., problems where we optimize other performance measures (e.g., TPR, FPR) or where models must obey constraints on model form or model predictions~\citep[e.g., group fairness constraints as in ][]{zafar2019fairness,celis2019classification,cotter2019optimization}. %

\paragraph{MIP Solvers}

We solve each MIP with a MIP solver such as CPLEX, CBC, or Gurobi. MIP solvers find the global optimum of a discrete optimization problem using exhaustive search algorithms like branch-and-bound~\citep[][]{wolsey1998integer}. In our setting, solving a MIP returns: (i) an upper bound on the objective value; (ii) a lower bound on the objective value; and (iii) the coefficients of a linear classifier that achieves the upper bound. When the upper bound matches the lower bound, the solution (iii) is \emph{certifiably optimal}, and our measures are exact. If a MIP solver does not return a certifiably optimal solution in a user-specified time limit, the bounds from (i) and (ii) can be used to produce bounds on ambiguity and discrepancy.

\subsection{Computing Discrepancy}
\label{Sec::MaximumDiscrepancy}

Given a training dataset, a baseline classifier $\baseclf{}$, and a user-specified error tolerance $\epsilon$, we compute the discrepancy over the $\epsilon$-level set around $\baseclf{}$ by solving the following optimization problem.
\begin{align}
\small
\begin{split}
\min_{h \in \Hset} \quad & \sum_{i=1}^n \indic{h(\xb_i) = \baseclf(\xb_i)} \\
\st \quad &\emprisk{h} \leq  \emprisk{\baseclf} + \epsilon
\end{split}\label{Opt::MaxDisc}
\end{align}
We denote the optimal solution to Equation \eqref{Opt::MaxDisc} as $\altclf{\epsilon}$. For linear classification problems, we can recover the coefficients of $\altclf{\epsilon}$ by solving the following MIP formulation, which we refer to as $\textsf{DiscMIP}(\baseclf{}, \epsilon)$:

\begin{subequations}
\label{IP::DiscIP}
\footnotesize
\begin{equationarray}{@{}c@{}r@{\,}c@{\,}l>{\,}l>{\,}r@{\;}}
\min_{} \quad & \sum_{i=0}^n a_i & & & & \notag \\
\st
& \quad M_i a_i  & \geq & \marginsym + \baseclf{}(\xb_i) \sum_{j=0}^d w_j x_{ij} & \miprange{i}{1}{n} & \label{Con::DiscrepancyIndicator} \\
& \epsilon  & \geq & \frac{1}{n} \sum_{i=1}^n y_i \baseclf{}(\xb_i) (1-a_i) & & \label{Con::DiscLevelSetConstraint} \\
& \coef{j} & = & \poscoef{j} + \negcoef{j} &  \miprange{j}{0}{d} & %
\label{Con::CoefficientsDisc} \\ 
& 1 & = & \sum_{j=0}^d (\poscoef{j} - \negcoef{j}) & & %
\label{Con::L1NormDisc} \\
& a_i & \in & \{0, 1\} & \miprange{i}{1}{n} & \notag \\
& \poscoef{j} & \in & [0,1] & \miprange{j}{0}{d} \notag \\
& \negcoef{j} & \in & [-1,0] & \miprange{j}{0}{d} \notag
\end{equationarray}
\end{subequations}

$\textsf{DiscMIP}$ minimizes the agreement between $h$ and $\baseclf{}$
using indicator variables $a_i = \indic{h(\xb_i) = \baseclf{}(\xb_i)}$. These variables are set via the ``Big-M'' constraints in \eqref{Con::DiscrepancyIndicator}. These constraints depend on: (i) a margin parameter $\gamma > 0$, which should be set to a small positive number (e.g., $\gamma = 10^{-4}$); and (ii) the Big-M parameters $M_i$, which can be set as $M_i = \marginsym + \max_{i}\Vert \xb_i \Vert_\infty$ since we have fixed $\|\coefs{}\|_1 = 1$ in constraint \eqref{Con::L1NormDisc}. Constraint \eqref{Con::DiscLevelSetConstraint} ensures that any feasible classifier must belong to the $\epsilon$-level set. %

\paragraph{Bounds}

Solving \textsf{DiscMIP} returns the coefficients of the classifier that maximizes discrepancy with respect to the baseline classifier $\baseclf{}$. If the solution is not certifiably optimal, the upper bound from \textsf{DiscMIP} corresponds to a lower bound on discrepancy. Likewise, the lower bound from \textsf{DiscMIP} corresponds to an upper bound on discrepancy.

\paragraph{Path Algorithm}

In Algorithm \ref{Alg::DiscrepancyPath}, we present a procedure to compute discrepancy for all possible values of $\epsilon$. The procedure solves $\textsf{DiscMIP}(\baseclf{}, \epsilon)$ for increasing values of $\epsilon \in \mathcal{E}$. At each iteration, it uses the current solution to initialize \textsf{DiscMIP} for the next iteration. The solution from the previous iteration produces upper and lower bounds that reduce the search space of the MIP, which is much faster than solving $\textsf{DiscMIP}$ separately for each $\epsilon$.

\begin{algorithm}[h]
\begin{algorithmic}[1]\small
\alginput{$\baseclf$}{baseline classifier}
\alginput{$\mathcal{E}$}{values of $\epsilon$ sorted in increasing order}
\For{$\epsilon \in \mathcal{E}$}
\State $\diffclf{\epsilon} \gets$ solution to $\textsf{DiscMIP}(h_0, \epsilon)$
\State $\singleConflicts \gets$ number of conflicts between $\diffclf{\epsilon}$ and $\baseclf{}$
\State $\epsilon_{\textrm{next}} \gets$ next value of $\epsilon \in \mathcal{E}$
\State Initialize $\textsf{DiscMIP}(h_0, \epsilon_{\textrm{next}})$ with $\diffclf{\epsilon}$
\label{AlgLine::DiscInitialization}
\EndFor{}
\Ensure $\{\singleConflicts, \diffclf{\epsilon}\}_{\epsilon \in \mathcal{E}}$ \hfill discrepancy and classifier for each $\epsilon$
\end{algorithmic}
\caption{Compute Discrepancy for All Values of $\epsilon$}
\label{Alg::DiscrepancyPath}
\end{algorithm}

\subsection{Computing Ambiguity}
\label{Sec::MeasuringAmbiguity}

We present an algorithm to compute ambiguity for all possible values of $\epsilon$. Given a baseline classifier $\baseclf{}$, the algorithm fits a \emph{pathological classifier} $\altclf{i}$ for each point in the training data -- i.e., the most accurate linear classifier that must assign a conflicting prediction to point $i$. Given a pathological classifier $\altclf{i}$ for each $i$, it then computes ambiguity over the $\epsilon$-level set by counting the number of pathological classifiers whose error is within $\epsilon$ of the error of the baseline classifier. Observe that ambiguity can be expressed as follows:
\begin{align*}
    \setConflicts(\baseclf{}) := &\frac{1}{n}\sum_{i=1}^{n} \max_{h \in \epsset{\baseclf{}}} \indic{h(\xb_i) \neq \baseclf{}(\xb_i)} \\ 
    = &\frac{1}{n} \sum_{i=1}^n \indic{\emprisk{\altclf{i}} \leq \emprisk{\baseclf{}} + \epsilon}.
\end{align*}
Thus, this approach corresponds to evaluating the summands in the expression for ambiguity in Definition \ref{Def::Ambiguity}. 

We fit $\altclf{i}$ by solving the following optimization problem:
\begin{align}
\small
\begin{split}
\min_{h \in \Hset} \quad & \sum_{i=1}^n \indic{h(\xb_i) \neq y_i} \\
\text{s.t.} \quad & h(\xb_i) \neq \baseclf{}(\xb_i)
\end{split}\label{Opt::Flipped}
\end{align}
Here, $h(\xb_i) \neq \baseclf{}(\xb_i)$ forces $h$ to assign a conflicting prediction to $\xb_i.$ For linear classification problems, we can recover the coefficients of $\altclf{i}$ by solving the following MIP formulation, which we refer to as $\textsf{FlipMIP}(h_0, \xb_i)$:

\vspace{-1em}
\begin{subequations}
\label{IP::FlipIP}
\footnotesize
\begin{equationarray}{@{}c@{}r@{\,}c@{\,}l>{\,}l>{\,}r@{\;}}
\min_{} & \quad \sum_{i=0}^n \error{i} & & & & \notag \\
\st 
& M_i \error{i} & \geq & y_i(\marginsym{} - \sum_{j=0}^d \coef{j} x_{ij}) &  \miprange{i}{1}{n}  & %
\label{Con::ErrorPositive} \\
 & \marginsym & \leq &  - \baseclf{}(\xb_i) \sum_{j=0}^d \coef{j} x_{j} 
 \label{Con::FlippedConstraint}\\
& \coef{j} & = & \poscoef{j} + \negcoef{j} &  \miprange{j}{0}{d} & %
\label{Con::Coefficients} \\ 
& 1 & = & \sum_{j=0}^d (\poscoef{j} - \negcoef{j}) & & %
\label{Con::L1Norm} \\
& \error{i} & \in & \{0,1\} & \miprange{i}{1}{n} \notag  \\
& \poscoef{j} & \in & [0,1] & \miprange{j}{0}{d} \notag \\
& \negcoef{j} & \in & [-1,0] & \miprange{j}{0}{d} \notag
\end{equationarray}
\end{subequations}
$\textsf{FlipMIP}$ minimizes the error rate of a pathological classifier $\altclf{i}$ using the indicator variables $\error{i} \gets \indic{h(\xb_i) \neq y_i}$. These variables are set through the Big-M constraints in \eqref{Con::ErrorPositive} whose parameters can be set in the same way as those in $\textsf{DiscMIP}$. Constraint \eqref{Con::FlippedConstraint} enforces the condition that $\altclf{i}(\xb) \neq \baseclf{}(\xb)$.

\paragraph{Bounds}

When a solver does not return a certifiably optimal solution to $\textsf{FlipMIP}$ within a user-specified time limit, it will return upper and lower bounds on the objective value of $\textsf{FlipMIP}$ that can be used to bound ambiguity. The upper bound will produce a lower bound on ambiguity. The lower bound will produce an upper bound on ambiguity.

\paragraph{Path Algorithm}

In Algorithm \ref{Alg::AmbiguityPath}, we present a procedure to efficiently compute ambiguity by initializing each instance of $\textsf{FlipMIP}$. In line \ref{Alg::AmbiguityPath}, the procedure sets the upper bound for $\textsf{FlipMIP}$ using the most accurate classifier in \textsf{POOL} that obeys the constraint $h(\xb_i) \neq \baseclf{}(\xb_i)$. Given a certifiably optimal baseline classifier, we can initialize the lower bound of $\textsf{FlipMIP}$ to $n\emprisk{\baseclf{}}$. 
\begin{algorithm}[h]
\begin{algorithmic}[1]\small
\alginput{$\baseclf$}{baseline classifier}
\alginput{$\mathcal{E}$}{values of $\epsilon$}
\alginitialize{$\textsf{POOL} \gets \emptyset$}{pool of pathological classifiers}
\For{$i \in \{1, 2, \dots, n\}$}
\State Initialize $\textsf{FlipMIP}(h_0, \xb_i)$ using best solution in $\textsf{POOL}$ \label{AlgLine::AmbiguityInitialize}
\State $\altclf{i} \gets$ solution to $\textsf{FlipMIP}(h_0, \xb_i)$ 
\State Add $\altclf{i}$ to $\textsf{POOL}$
\EndFor{}
\For{$\epsilon \in \mathcal{E}$}
\State $\setConflicts \gets \frac{1}{n} \sum_{i=1}^n \indic{\emprisk{\altclf{i}} \leq \emprisk{\baseclf{}}}$
\EndFor{}%
\Ensure $\{\setConflicts\}_{\epsilon \in \mathcal{E}}$ and $\{\altclf{i}\}_{i=1}^n$
\end{algorithmic}
\caption{Compute Ambiguity for All Values of $\epsilon$}
\label{Alg::AmbiguityPath}
\end{algorithm}

\section{Experiments}
\label{Sec::Experiments}

In this section, we apply our tools to measure predictive multiplicity in recidivism prediction problems. We have three goals: (i) to measure the incidence of predictive multiplicity in real-world classification problems; (ii) to discuss how reporting predictive multiplicity can inform stakeholders; (iii) to show that we can also measure predictive multiplicity using existing tools, albeit imperfectly. We include software to reproduce our results at \gitrepo{}.

Our focus on recidivism prediction should \emph{not} be viewed as an endorsement of the practice. We consider recidivism prediction since it is a domain where predictive multiplicity has serious ethical implications, and where the existence of predictive multiplicity may serve as an additional reason to forgo the deployment of machine learning entirely~\citep[see e.g.,][for broader critiques]{harcourt2008against,lum2016predict,barabas2017interventions}. 

\subsection{Setup}

\paragraph{Datasets}

We derive 8 datasets from the following studies of recidivism in the United States: 
\begin{itemize}
\item \textds{compas} from \citealt{angwin2016machine}; %
\item \textds{pretrial} from \hyperlink{https://www.icpsr.umich.edu/icpsrweb/NACJD/studies/2038}{Felony Defendants in Large Urban Counties}~\citep{doj2009recidivism}; 
\item \textds{recidivism} from \hyperlink{https://www.icpsr.umich.edu/icpsrweb/NACJD/studies/3355/version/8}{Recidivism of Prisoners Released in 1994}~\citep{doj1994recidivismdata}. %
\end{itemize}
We process each dataset by binarizing features and dropping examples with missing entries. For clarity of exposition, we oversample the minority class to equalize the number of positive and negative examples. Oversampling allows us to report our measures for level sets defined in terms of error rates instead of TPR/FPR. We find that oversampling has a negligible effect on our measures of multiplicity.  We provide a summary of each datasets in Table \ref{Table::BaselineComparisons}. %

\newcommand{\datacell}[3]{{\cell{l}{\small\texttt{#1}\\$n={#2}$\\$d={#3}$}}}

\paragraph{Measurement Protocol}

We compute our measures of predictive multiplicity for each dataset as follows. We split each dataset into a \emph{training set} composed of $80\%$ of points and a \emph{test set} composed of 20\% of points. We use the training set to fit a \emph{baseline classifier} that minimizes the 0-1 loss directly by solving MIP \eqref{IP::BaselineIP} in Appendix \ref{Appendix::MIP}.  We measure ambiguity and discrepancy for \emph{all possible values} of the error tolerance $\epsilon$ using Algorithms \ref{Alg::DiscrepancyPath} and \ref{Alg::AmbiguityPath}. We solve each MIP on a 3.33 GHz CPU with 16 GB RAM. We allocate at most 6 hours to fit the baseline model, 6 hours to fit the models to compute discrepancy for all $\epsilon$, and 6 hours to fit the models to compute ambiguity for all $\epsilon$.

\begin{table}[tb]
\resizebox{1.0\linewidth}{!}{
\begin{tabular}{llrrrr}
\toprule
& & & &
\multicolumn{2}{c}{Error of $\baseclf{}$} \\ 
\cmidrule(lr){5-6} 
\cell{l}{Dataset} & 
\cell{l}{Outcome Variable} &
\cell{r}{\large{$n$}} & 
\cell{r}{\large{$d$}} &    
\cell{c}{Train}& 
\cell{c}{Test}
\\
\midrule

\textds{compas\_arrest} &         
rearrest for any crime &
5,380 & 18 &
32.7\% &      33.4\% \\

\textds{\small{compas\_violent}} & 
rearrest for violent crime &
8,768 & 18 &
 37.7\% &       37.9\% \\
 
\textds{pretrial\_CA\_arrest} &  
 rearrest for any crime &
9,926 & 22 &
34.1\%  &  34.4\% \\

\textds{pretrial\_CA\_fta} &           
failure to appear &
8,738 & 22 &
36.3\% &      36.3\%  \\

\textds{recidivism\_CA\_arrest} &  
rearrest for any offense &
114,522 & 20 &
34.4\% &      34.2\%  \\

\textds{recidivism\_CA\_drug} &  
rearrest for drug-related offense &
96,664 & 20 &
36.3\% &      36.2\%  \\

 \textds{recidivism\_NY\_arrest} &  
 rearrest for any offense &
 31,624 & 20 &
31.0\% & 31.8\% \\

\textds{recidivism\_NY\_drug} & 
rearrest for drug-related offense &
27,526 & 20 &
32.5\% &  33.6\%  \\

\bottomrule
\end{tabular}
}
\caption{Recidivism prediction datasets used in Section \ref{Sec::Experiments}. For each dataset, we fit a baseline linear classifier that minimizes training error. As shown, the models generalize as training error is close to test error. This is expected given that we fit models from a simple hypothesis class. Here, $n$ and $d$ denote the number of examples and features in each dataset, respectively. All datasets are publicly available. We include a copy of \textds{compas\_arrest} and \textds{compas\_violent} with our code. The remaining datasets must be requested from ICPSR due to privacy restrictions.}
\label{Table::BaselineComparisons}
\end{table}

\newcommand{\coeftbl}[3]{\resizebox{0.15\linewidth}{!}{\input{figure/#1_#2_model_#3}}}
\newcommand{\comparisonTable}[2]{\resizebox{0.9\linewidth}{!}{\input{figure/#1_#2_comparison_table.tex}}}

\newcommand{\getDiscPath}[1]{figure/#1_disc_vs_epsilon_path.png}
\newcommand{\getDiscPathGlmnet}[1]{figure/#1_glmnet_disc_plot.png}
\newcommand{\getFlippedPath}[1]{figure/#1_flipped_error_path_duplicated_transpose.png}
\newcommand{\getFlippedPathGlmnet}[1]{figure/#1_glmnet_flipped_plot.png}

\newcommand{\errorprofilerow}[2]{%
{\scriptsize\sccell{l}{#1}}&%
\cell{c}{\includegraphics[width=0.35\linewidth]{\getDiscPath{#2}}}&%
\cell{c}{\includegraphics[width=0.35\linewidth]{\getFlippedPath{#2}}}%
}

\newcommand{\errorprofilerowglmnet}[2]{%
{\scriptsize\sccell{l}{#1}}&%
\cell{c}{\includegraphics[width=0.35\linewidth]{\getDiscPathGlmnet{#2}}}&%
\cell{c}{\includegraphics[width=0.35\linewidth]{\getFlippedPathGlmnet{#2}}}%
}

\newcommand{\errorprofilerowPlaceholder}[2]{%
{\scriptsize\ttfamily\cell{c}{#1}}&%
\cell{c}{\includegraphics[width=0.45\linewidth]{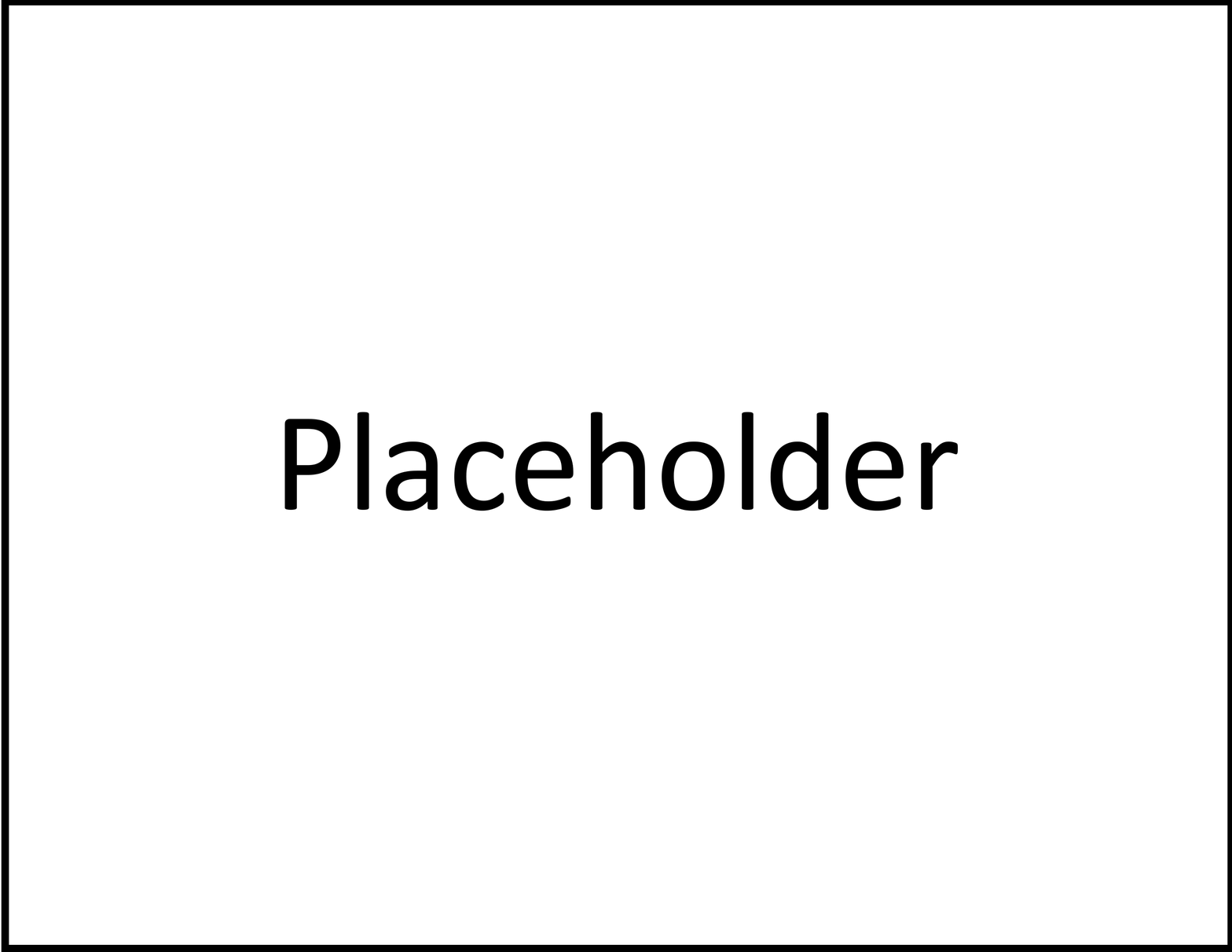}}&%
\cell{c}{\includegraphics[width=0.45\linewidth]{figure/placeholder.pdf}}%
}

\begin{figure*}[htbp]
    \centering
    \scriptsize
    \resizebox{\linewidth}{!}{
    \begin{tabular}{ccc}
    \toprule
    \sccell{l}{Approach} & 
    \sccell{c}{Discrepancy} & 
    \sccell{c}{Ambiguity}   \\ 
    \toprule
    \errorprofilerow{Exact}{compas_arrest_small} \\ 
    \midrule
    \errorprofilerowglmnet{Ad Hoc}{compas_arrest_small} \\ 
    \bottomrule
    \end{tabular}
    }
    \caption{Severity of predictive multiplicity measured using our tools (top) and using an ad hoc approach (bottom) for \textds{compas\_arrest}. We plot the values of discrepancy (left) and ambiguity (right) over the $\epsilon$-level set. We find a discrepancy of 17\% and ambiguity of $44\%$ over the $1\%$-level set. This means that one can change $17\%$ of predictions by switching the baseline model with a model that is only 1\% less accurate, and that $44\%$ of individuals are assigned conflicting predictions by models in the 1\%-level set. We include similar plots for other datasets in Appendix \ref{Appendix::Experiments}. %
    }
    \label{Fig::ErrorProfiles}
\end{figure*}

\renewcommand{\errorprofilerow}[2]{%
{\scriptsize\ttfamily\cell{c}{#1}}&%
\cell{c}{\includegraphics[width=0.35\linewidth]{\getDiscPath{#2}}}&%
\cell{c}{\includegraphics[width=0.35\linewidth]{\getFlippedPath{#2}}}%
}

\renewcommand{\errorprofilerowglmnet}[2]{%
{\scriptsize\ttfamily\cell{c}{#1}}&%
\cell{c}{\includegraphics[width=0.35\linewidth]{\getDiscPathGlmnet{#2}}}&%
\cell{c}{\includegraphics[width=0.35\linewidth]{\getFlippedPathGlmnet{#2}}}%
}

\begin{table*}[t]
\centering
\resizebox{0.9\linewidth}{!}{\scriptsize
\begin{tabular}{llll}

\toprule
&
\bfcell{c}{Baseline Model} &
\bfcell{c}{Individual Ambiguity Model} &
\bfcell{c}{Discrepancy Model} \\

\toprule
$h(\xb_p)$  & $+1$ & $-$1 & $-$1 \\
\toprule

Error (Train/Test) 
& 32.7\% / 33.4\%  
& 32.7\% / 33.4\% &  
33.6 / 34.5\% \\

\toprule

Discrepancy (Train/Test) & 
0.0\% / 0.0\% & 
0.0037\% / 0.0\% & 
16.8\% / 15.1\% \\ 


\toprule
\cell{l}{Score\\Function}     &  
\resizebox{0.15\linewidth}{!}{\begin{tabular}{@{\;}c@{\;}r@{\,}l@{\;}c@{\;}r@{\,}l}
 $+$ &   0.5 &                       \textfn{age$\leq$25} \\  $+$ &   0.0 &                        \textfn{age\_25-45} \\
 $-$ &  16.4 &                       \textfn{age$\geq$46} \\  $-$ &  16.3 &                            \textfn{female} \\
 $-$ &   0.2 &                     \textfn{n\_priors$=$0} \\  $-$ &   0.1 &                  \textfn{n\_priors$\geq$1} \\
 $+$ &  16.4 &                  \textfn{n\_priors$\geq$2} \\  $+$ &  16.6 &                  \textfn{n\_priors$\geq$5} \\
 $+$ &   0.0 &     \textfn{n\_juvenile\_misdemeanors$=$0} \\  $-$ &   0.1 &  \textfn{n\_juvenile\_misdemeanors$\geq$1} \\
 $+$ &   0.0 &  \textfn{n\_juvenile\_misdemeanors$\geq$2} \\  $-$ &  32.6 &  \textfn{n\_juvenile\_misdemeanors$\geq$5} \\
 $+$ &   0.0 &         \textfn{n\_juvenile\_felonies$=$0} \\  $-$ &   0.2 &      \textfn{n\_juvenile\_felonies$\geq$1} \\
 $+$ &   0.3 &      \textfn{n\_juvenile\_felonies$\geq$2} \\  $+$ &   0.0 &      \textfn{n\_juvenile\_felonies$\geq$5} \\
 $-$ &   0.2 &                \textfn{charge\_degree$=$M} \\  $+$ &   0.0 &                                            \\
     &       &                                            &      &       &                                            \\
\end{tabular}
} &  
\resizebox{0.15\linewidth}{!}{\begin{tabular}{@{\;}c@{\;}r@{\,}l@{\;}c@{\;}r@{\,}l}
 $+$ &  10.3 &                       \textfn{age$\leq$25} \\  $+$ &   0.0 &                        \textfn{age\_25-45} \\
 $-$ &   9.9 &                       \textfn{age$\geq$46} \\  $-$ &   9.7 &                            \textfn{female} \\
 $+$ &   0.0 &                     \textfn{n\_priors$=$0} \\  $+$ &   0.0 &                  \textfn{n\_priors$\geq$1} \\
 $+$ &  19.8 &                  \textfn{n\_priors$\geq$2} \\  $+$ &  10.1 &                  \textfn{n\_priors$\geq$5} \\
 $+$ &   0.0 &     \textfn{n\_juvenile\_misdemeanors$=$0} \\  $-$ &   0.1 &  \textfn{n\_juvenile\_misdemeanors$\geq$1} \\
 $-$ &  10.1 &  \textfn{n\_juvenile\_misdemeanors$\geq$2} \\  $-$ &   9.5 &  \textfn{n\_juvenile\_misdemeanors$\geq$5} \\
 $-$ &   9.9 &         \textfn{n\_juvenile\_felonies$=$0} \\  $-$ &  10.1 &      \textfn{n\_juvenile\_felonies$\geq$1} \\
 $+$ &   0.3 &      \textfn{n\_juvenile\_felonies$\geq$2} \\  $+$ &   0.0 &      \textfn{n\_juvenile\_felonies$\geq$5} \\
 $-$ &   0.2 &                \textfn{charge\_degree$=$M} \\  $+$ &   0.0 &                                            \\
     &       &                                            &      &       &                                            \\
\end{tabular}
} &
\resizebox{0.15\linewidth}{!}{\begin{tabular}{@{\;}c@{\;}r@{\,}l@{\;}c@{\;}r@{\,}l}
 $+$ &   7.7 &                       \textfn{age$\leq$25} \\  $+$ &   0.0 &                        \textfn{age\_25-45} \\
 $-$ &   7.8 &                       \textfn{age$\geq$46} \\  $-$ &   7.6 &                            \textfn{female} \\
 $-$ &   7.8 &                     \textfn{n\_priors$=$0} \\  $+$ &   0.0 &                  \textfn{n\_priors$\geq$1} \\
 $+$ &   7.4 &                  \textfn{n\_priors$\geq$2} \\  $+$ &   7.8 &                  \textfn{n\_priors$\geq$5} \\
 $+$ &   0.0 &     \textfn{n\_juvenile\_misdemeanors$=$0} \\  $+$ &   0.1 &  \textfn{n\_juvenile\_misdemeanors$\geq$1} \\
 $-$ &   0.1 &  \textfn{n\_juvenile\_misdemeanors$\geq$2} \\  $-$ &  15.2 &  \textfn{n\_juvenile\_misdemeanors$\geq$5} \\
 $+$ &   7.7 &         \textfn{n\_juvenile\_felonies$=$0} \\  $+$ &   0.0 &      \textfn{n\_juvenile\_felonies$\geq$1} \\
 $+$ &  15.4 &      \textfn{n\_juvenile\_felonies$\geq$2} \\  $+$ &   0.0 &      \textfn{n\_juvenile\_felonies$\geq$5} \\
 $-$ &   7.5 &                \textfn{charge\_degree$=$M} \\  $-$ &   0.1 &                                            \\
     &       &                                            &      &       &                                            \\
\end{tabular}
} \\



\bottomrule
\end{tabular}
}
\caption{Competing linear classifiers that assign conflicting prediction to $\xb_p$  \textds{compas\_arrest}. We show the baseline model (left), the competing model fit to measure ambiguity to $\xb_p$ (middle), and competing model fit to measure discrepancy (right). The baseline model predicts $h(\xb_p) = +1$ while other models predict $h(\xb_p) = -1$. 
As shown, there exists at least two competing models that predict that $\xb_p$ would not recidivate. In addition, each model exhibits different coefficients and measures of variable importance.
}
\label{Table::ModelComparisons}
\end{table*}

\paragraph{Ad Hoc Measurement Protocol}

We compute ambiguity and discrepancy through an ad hoc approach. We include these results to show that an imperfect analysis of predictive multiplicity can reveal salient information. Here, we produce a pool of competing models using the \textsf{glmnet} package of \citet{friedman2010glmnet}. We fit 1,100 linear classifiers using penalized logistic regression. Each model corresponds to an optimizer of the logistic loss with a different degree of $\ell_1$ and $\ell_2$ regularization. We choose the baseline model as the model that minimizes the 5-fold CV test error.

\subsection{Results}

In Figure \ref{Fig::ErrorProfiles}, we plot ambiguity and discrepancy for all possible values of the error tolerance $\epsilon$ for \textds{compas\_arrest}, comparing the measures produced using our tools to those produced using an ad hoc analysis. In Table \ref{Table::ModelComparisons}, we compare competing classifiers for \textds{compas\_arrest}. In what follows, we discuss these results. %

\paragraph{On the Incidence of Predictive Multiplicity}

Our results in Figure \ref{Fig::ErrorProfiles} show how predictive multiplicity arises in real-world prediction problems. For the 8 datasets we consider, we find that between $4\%$ and $53\%$ of individuals are assigned conflicting predictions in the 1\%-level set. In \textds{compas\_arrest}, for example, we observe an ambiguity of 44\%. %
Considering discrepancy, we can find a competing model in the $1\%$-level set that would assign a conflicting prediction to $17\%$ of individuals. %

\paragraph{On the Burden of Multiplicity}

Our results show that the incidence of multiplicity can differ significantly between protected groups. In \textds{compas\_violent}, for example, predictive multiplicity disproportionately affects African-Americans compared to individuals of other ethnic groups: the proportion of individuals who are assigned conflicting predictions over the $1\%$ level set is $72.9\%$ for African-Americans but $37.2\%$ for Caucasians. Groups with a larger burden of multiplicity are more vulnerable to model selection, and more likely to be affected by the ignorance of competing models.

\paragraph{On the Implications of Predictive Multiplicity}

Our results illustrate how reporting ambiguity and discrepancy can challenge model development and deployment. In \textds{compas\_arrest}, for example, our baseline model provably optimizes training error and generalizes. Without an analysis of predictive multiplicity, practitioners could deploy this model. Our analysis reveals that there exists a competing model that assigns conflicting predictions to $17\%$ of individuals. Thus, these measures support the need for greater scrutiny and stakeholder involvement in model selection.

Reporting ambiguity and discrepancy also help us calibrate trust in downstream processes in the modern machine learning life-cycle \citep[e.g., evaluating feature influence as in][]{kumar2020problems,marx2019disentangling}. Consider the process of explaining individual predictions. In this case, an ambiguity of 44\% means one could produce conflicting explanations for 44\% of predictions. While every explanation would help us understand how competing models operate, evidence of conflicting predictions would provide a safeguard against unwarranted rationalization.

\paragraph{On Model Selection}

When presented with many competing models, a natural solution is to choose among them to optimize secondary objectives. We support this practice when secondary objectives reflect bona fide goals rather than a way to resolve reconciling multiplicity (see Section \ref{Sec::BroaderDiscussion} for a discussion). However, tie-breaking does not always yield a unique model.  For example, on the \textds{compas\_arrest} dataset, we can break ties between competing models in the 1\%-level set on the basis of a group fairness criterion (i.e., by minimizing the disparity in accuracy between African-Americans and other ethic groups). In this case, we find 102 competing models that are also within 1\% optimal in terms of the secondary criterion.

\paragraph{On Ad Hoc Measurement}

Our results for the ad hoc approach show how measuring and reporting predictive multiplicity can reveal useful information even without specialized tools. In \textds{compas\_arrest}, for example, an ad hoc analysis reveals an ambiguity of 10\% and a discrepancy of 7\% over the set of competing models. These estimates are far less than those produced using our tools ($44\%$ and $17\%$ respectively). This is because the ad hoc approach only considers competing models that can be obtained by varying $\ell_1$ and $\ell_2$ penalties in penalized logistic regression, rather than all linear classifiers in the $1\%$-level set. These results show that ad hoc approaches can detect predictive multiplicity, but should not be used to certify the absence of multiplicity.

\section{Concluding Remarks}
\label{Sec::BroaderDiscussion}

Prediction problems can exhibit predictive multiplicity due to a host of reasons, including feature selection, a misspecified hypothesis class, or the existence of latent groups. 

Even as there exist techniques to choose between competing models, we do not advocate a general prescription to resolve predictive multiplicity. Instead, we argue that we should measure and report multiplicity like we measure and report test error \citep[][]{saleiro2018aequitas,reisman2018algorithmic}. In this way, predictive multiplicity can be resolved on a case-by-case basis, and in a way that allows for input from stakeholders \citep[as per the principles of contestable design; see e.g.,][]{hirsch2017designing,kluttz2018contestability}.

Reporting predictive multiplicity can change how we build and deploy models in human-facing applications. In such settings, presenting stakeholders with meaningful information about predictive multiplicity may lead them to think carefully about which model to deploy, consider assigning favorable predictions to individuals who receive conflicting predictions, or forgo deployment entirely.

\section*{Acknowledgements}
We thank Sorelle Friedler, Dylan Slack, and Ben Green for helpful discussions, and anonymous reviewers for constructive feedback. This research is supported in part by the National Science Foundation under Grants No. CAREER CIF-1845852 and by a Google Faculty Award.

\bibliographystyle{icml2020}
\bibliography{predictive_multiplicity}

\appendix
\onecolumn

\section{Omitted Proofs}
\label{Appendix::Proofs}

\paragraph{Proof of Proposition \ref{Prop::maxConflictsBound}}

We use the Triangle Inequality to bound the distance between the vector of predictions of the baseline model and the predictions of a competing model in the $\epsilon$-level set. Let 
$y=\{y_i\}_{i=1}^n$
be the vector of labels, let
$\hat{y}=\{\baseclf(\xb_i)\}_{i=1}^n$
be the vector of predictions of the baseline model, and let $y'=\{h'(\xb_i)\}_{i=1}^n$ be the predictions of a competing model $h'$ in the $\epsilon$-level set. Note that 
$y,  y', \hat{y} \in \{+1, -1\}^n$. 
Now, we can express the risk of the baseline model $\emprisk{\baseclf{}}$, the risk of the competing model $\emprisk{h'}$, and the discrepancy between $h$ and $h'$ ,denoted $\cfun{\baseclf{}}{h'}$, in terms of these three vectors by 
\begin{align*}
\emprisk{\baseclf{}} &= \frac{1}{4}\Vert y - \hat{y} \Vert \\ \emprisk{h'} &= \frac{1}{4}\Vert y - y' \Vert \\
\cfun{\baseclf{}}{h'} &= \frac{1}{4}\Vert y' - \hat{y} \Vert
\end{align*}
Next, consider the triangle formed in $\mathbb{R}^n$ by the points $y, y'$ and $\hat{y}$, with side lengths $\Vert y - \hat{y} \Vert$, $\Vert y' - \hat{y} \Vert$ and $\Vert y - y' \Vert$. The Triangle Inequality gives us that 
$$\Vert y' - \hat{y} \Vert \leq \Vert y - y' \Vert + \Vert y - \hat{y} \Vert.$$ 
Substituting using the three equations above, we have
$$\cfun{\baseclf}{h'} \leq \emprisk{\baseclf{}} + \emprisk{h'}.$$
Since $h' \in \epsset{\baseclf{}}$, we have by the definition of the $\epsilon$-level set that 
$\emprisk{h'} \leq \emprisk{\baseclf{}} + \epsilon$.
We can then rewrite the above expression to yield
$$\cfun{\baseclf{}}{h'} \leq 2\emprisk{\baseclf{}} + \epsilon$$
Recall that $\singleConflicts(h_0) := \max_{h' \in \epsset{\baseclf{}}} \cfun{\baseclf{}}{h'}$. Since each $h' \in \epsset{\baseclf{}}$ satisfies $\cfun{\baseclf{}}{h'} \leq 2\emprisk{\baseclf{}} + \epsilon$, we have the result that $\singleConflicts(h_0) \leq 2\emprisk{\baseclf{}} + \epsilon$. \hfill $\square$

\newpage
\section{MIP Formulation for Training the Best Linear Classifier}
\label{Appendix::MIP}

We fit a classifier that minimizes the training error by solving an optimization problem of the form:
\begin{align}
\small
\min_{h \in \Hset} \quad \sum_{i=1}^n \indic{h(\xb_i) \neq y_i} \label{Opt::ZeroOne}
\end{align}

We solve this optimization problem via the following MIP formulation:

\begin{subequations}
\label{IP::BaselineIP}
\begin{equationarray}{@{}c@{}r@{\,}c@{\,}l>{\,}l>{\,}r@{\;}}
\min_{} & \quad \sum_{i=0}^n \error{i} & & & & \notag \\
\st 
& M_i \error{i} & \geq & y_i(\marginsym{} - \sum_{j=0}^d \coef{j} x_{ij}) &  \miprange{i}{1}{n}  & %
\label{Con::ErrorPositiveX} \\
 & \marginsym & \leq &  - \baseclf(\xb_j) \sum_{j=0}^d \coef{j} x_{ij} 
 \label{Con::FlippedConstraintX}\\
& 1 & = &  \error{i} + \error{i'}  &  (i, i') \in \conflictset{} & %
\label{Con::ConflictingPointsX} \\
& \coef{j} & = & \poscoef{j} + \negcoef{j} &  \miprange{j}{0}{d} & %
\label{Con::CoefficientsX} \\ 
& 1 & = & \sum_{j=0}^d (\poscoef{j} - \negcoef{j}) & & %
\label{Con::L1NormX} \\
& \error{i} & \in & \{0,1\} & \miprange{i}{1}{n} \notag  \\
& \coef{j} & \in & [-1,1] & \miprange{j}{0}{d} \notag \\
& \poscoef{j} & \in & [0,1] & \miprange{j}{0}{d} \notag \\
& \negcoef{j} & \in & [-1,0] & \miprange{j}{0}{d} \notag
\end{equationarray}
\end{subequations}

Here, constraints \eqref{Con::ErrorPositiveX} set the mistake indicators $\error{i} \gets \indic{h(\xb_i) \neq y_i}.$ These constraints depend on: (i) a margin parameter $\gamma > 0$, which should be set to a small positive number (e.g., $\gamma = 10^{-4}$); and (ii) the ``Big-M" parameters $M_i$ which can be set as $M_i = \marginsym + \max_{\xb_i \in X}\Vert \xb_i \Vert_\infty$ since we have fixed $\|\coefs{}\|_1 = 1$ in constraint \eqref{Con::L1NormX}. 
Constraint \eqref{Con::ConflictingPointsX} produces an improved lower bound by encoding the necessary condition that any classifier must make exactly one mistake between any two points $(i,i') \in \conflictset{}$ with identical features $\xb_i = \xb_{i'}$ and conflicting labels. Here, $\conflictset{} = \{(i, i'): \xb_i = \xb_{i'}, y_i = +1, y_{i'} = -1\}$ is the set of points with conflicting labels.

\newpage
\section{Additional Experimental Results}
\label{Appendix::Experiments}

\begin{figure*}[h]
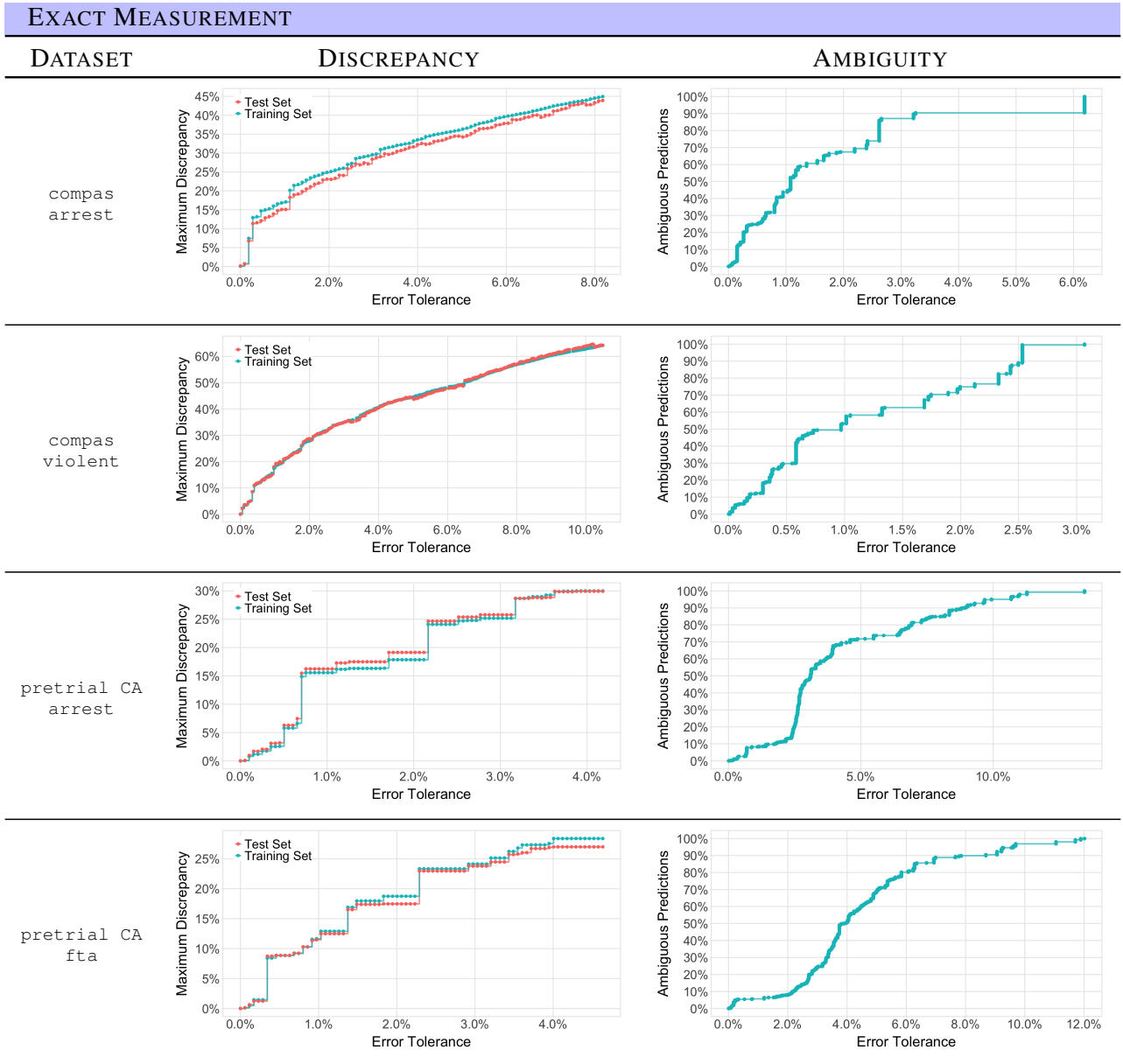

    \centering
    \resizebox{\linewidth}{!}{
    \begin{tabular}{ccc}
    \multicolumn{3}{l}{\cellcolor{blue!25} \textsc{Exact Measurement}} \\
        
    \toprule
    \sccell{c}{Dataset} & 
    \sccell{c}{Discrepancy} & 
    \sccell{c}{Ambiguity} \\ 
    \toprule
    \errorprofilerow{compas\\arrest}{compas_arrest_small} \\  \midrule
    \errorprofilerow{compas\\violent}{compas_violent_small} \\  \midrule
    \errorprofilerow{pretrial CA\\arrest}{pretrial_CA_arrest} \\  \midrule
    \errorprofilerow{pretrial CA\\fta}{pretrial_CA_fta} \\
    \bottomrule
    \end{tabular}
    }
\caption{Multiplicity profiles for the \textds{compas} and \textds{pretrial} datasets.}
\label{Fig::SuppMIP1}
\end{figure*}

\begin{figure*}[h]
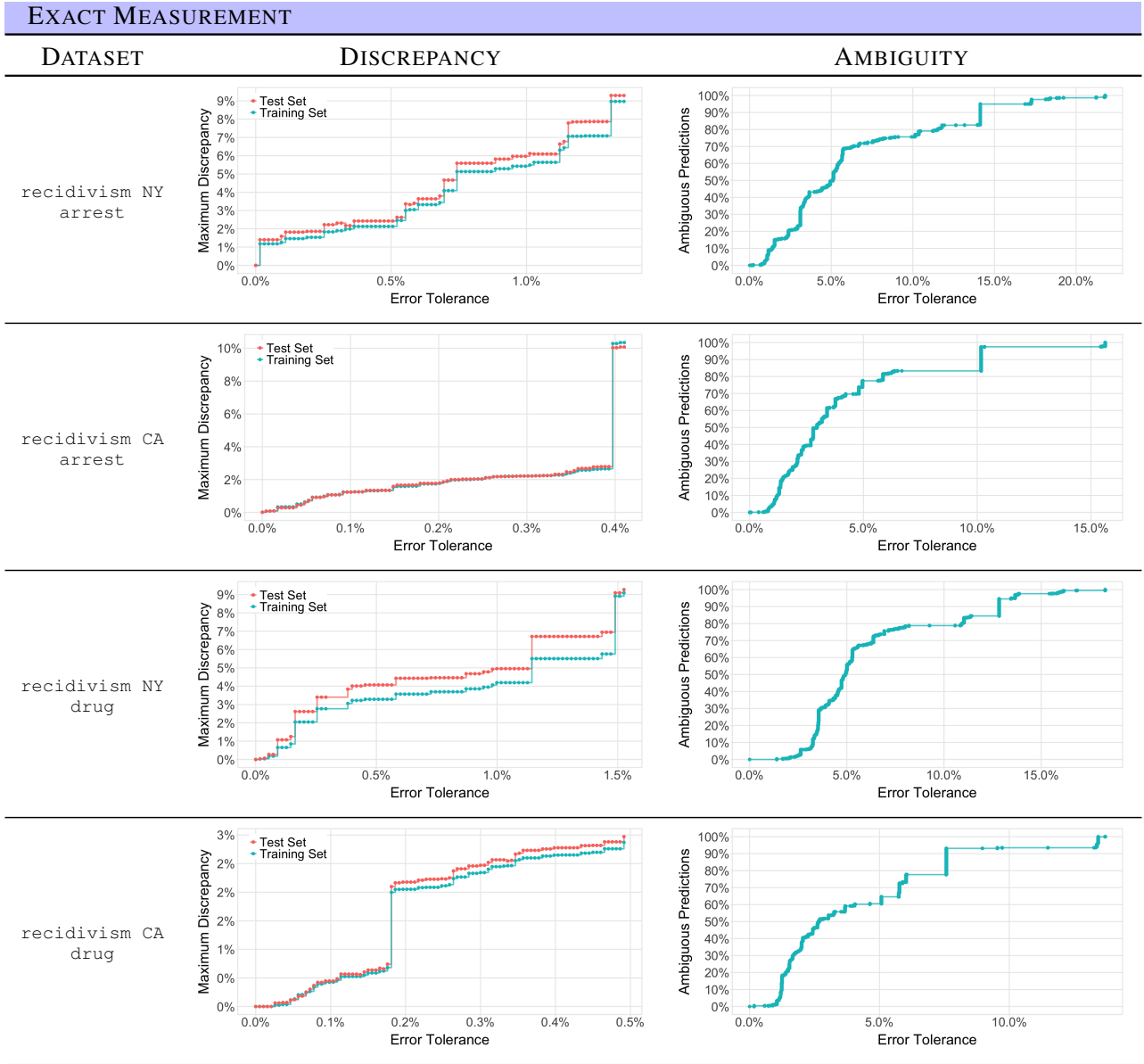

    \centering
    \resizebox{\linewidth}{!}{
    \begin{tabular}{ccc}
    \multicolumn{3}{l}{\cellcolor{blue!25} \textsc{Exact Measurement}} \\ 
    \toprule
    \sccell{c}{Dataset} & 
    \sccell{c}{Discrepancy} & 
    \sccell{c}{Ambiguity} \\ 
    \toprule
    \errorprofilerow{recidivism NY\\arrest}{recidivism_NY_arrest} \\  \midrule
    \errorprofilerow{recidivism CA\\arrest}{recidivism_CA_arrest} \\ \midrule
    \errorprofilerow{recidivism NY\\drug}{recidivism_NY_drug} \\ \midrule
    \errorprofilerow{recidivism CA\\drug}{recidivism_CA_drug}
    \\ 
    \bottomrule
    \end{tabular}
    }
    \caption{Multiplicity profiles for the \textds{recidivism} datasets.}
\label{Fig::SuppMIP2}
\end{figure*}

\begin{figure*}[h]
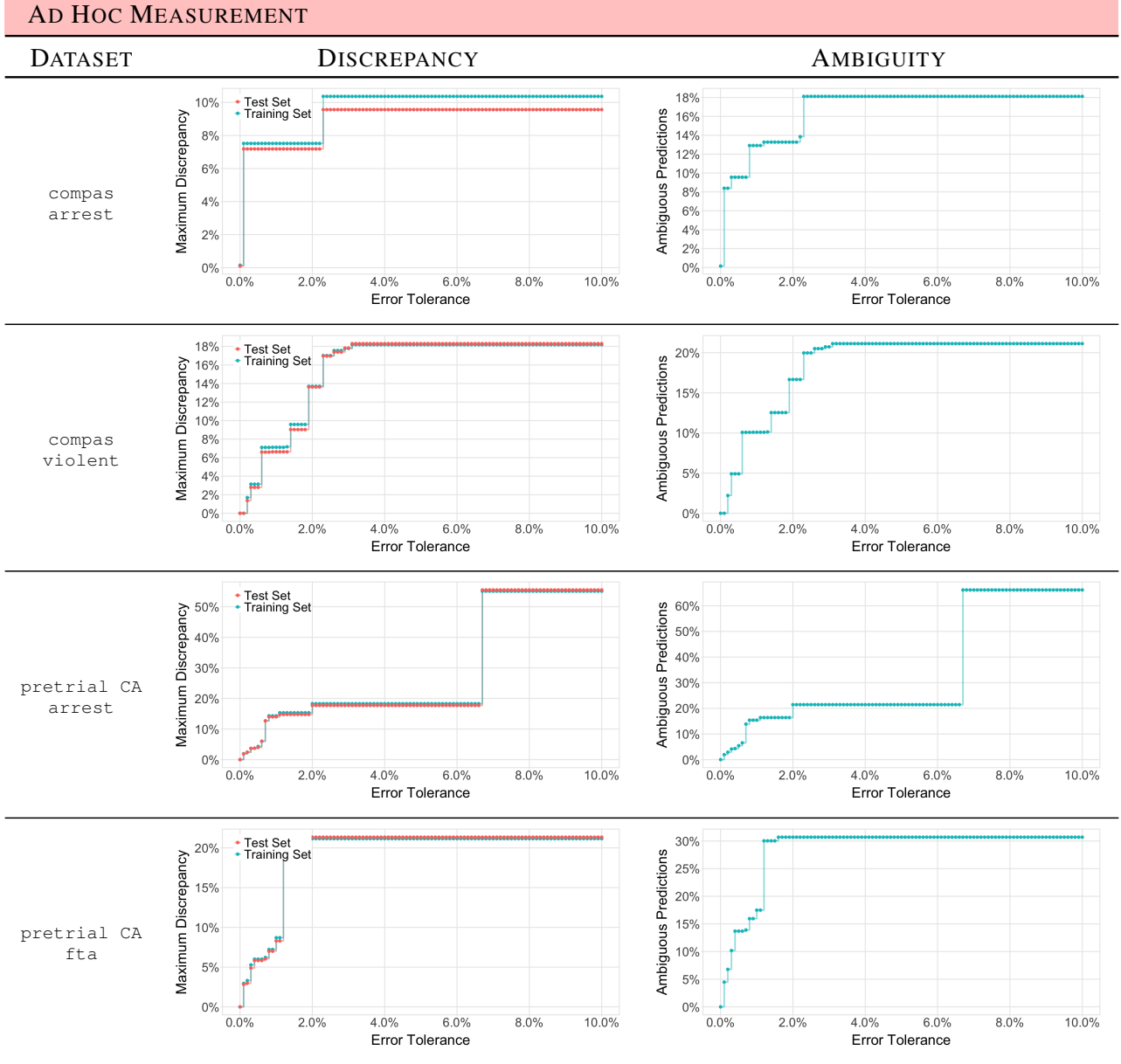

    \centering
    \resizebox{\linewidth}{!}{
    \begin{tabular}{ccc}
    \multicolumn{3}{l}{\cellcolor{red!25} \textsc{Ad Hoc Measurement}} \\[0.1em]
    \toprule
    \sccell{c}{Dataset} & 
    \sccell{c}{Discrepancy} & 
    \sccell{c}{Ambiguity} \\ 
    \toprule
    \errorprofilerowglmnet{compas\\arrest}{compas_arrest_small} \\  \midrule
    \errorprofilerowglmnet{compas\\violent}{compas_violent_small} \\  \midrule
    \errorprofilerowglmnet{pretrial CA\\arrest}{pretrial_CA_arrest} \\  \midrule
    \errorprofilerowglmnet{pretrial CA\\fta}{pretrial_CA_fta} \\  %
    \bottomrule
    \end{tabular}
    }
    \caption{Multiplicity profiles for the \textds{compas} and \textds{pretrial} datasets produced via pools of logistic regression models.}
\label{Fig::SuppGlmnet1}
\end{figure*}

\begin{figure*}[h]
    \centering
    \resizebox{\linewidth}{!}{
    \begin{tabular}{ccc}
    \multicolumn{3}{l}{\cellcolor{red!25} \textsc{Ad Hoc Measurement}} \\[0.1em]
    \toprule
    \sccell{c}{Dataset} & 
    \sccell{c}{Discrepancy} & 
    \sccell{c}{Ambiguity} \\ 
    \toprule
    \errorprofilerowglmnet{recidivism NY\\arrest}{recidivism_NY_arrest} \\  \midrule
    \errorprofilerowglmnet{recidivism CA\\arrest}{recidivism_CA_arrest} \\ \midrule
    \errorprofilerowglmnet{recidivism NY\\drug}{recidivism_NY_drug} \\ \midrule
    \errorprofilerowglmnet{recidivism CA\\drug}{recidivism_CA_drug} \\
    \bottomrule
    \end{tabular}
    }
    \caption{Multiplicity profiles for the \texttt{recidivism} datasets produced via pools of logistic regression models.}
\label{Fig::SuppGlmnet2}
\end{figure*}

\end{document}